# AVDNet: A Small-Sized Vehicle Detection Network for Aerial Visual Data

Murari Mandal, Manal Shah, Prashant Meena, Sanhita Devi, and Santosh Kumar Vipparthi

*Abstract*—Detection of small-sized targets in aerial views is a challenging task due to the smallness of vehicle size, complex background, and monotonic object appearances. In this letter, we propose a one-stage vehicle detection network (AVDNet) to robustly detect small-sized vehicles in aerial scenes. In AVDNet, we introduced ConvRes residual blocks at multiple scales to alleviate the problem of vanishing features for smaller objects caused because of the inclusion of deeper convolutional layers. These residual blocks, along with enlarged output feature map, ensure the robust representation of the salient features for small-sized objects. Furthermore, we proposed a recurrent-feature aware visualization (RFAV) technique to analyze the network behavior. We also created a new airborne image data set (ABD) by annotating 1396 new objects in 79 aerial images for our experiments. The effectiveness of AVDNet is validated on VEDAI, DLR-3K, DOTA, and the combined (VEDAI, DLR-3K, DOTA, and ABD) data set. Experimental results demonstrate the significant performance improvement of the proposed method over state-of-the-art detection techniques in terms of mAP, computation, and space complexity.

## I. INTRODUCTION

ADVANCES in unmanned aerial vehicles (UAVs) technology has unlocked a new frontier of computer vision, which requires analysis and interpretation of aerial images and videos. Vehicle detection in aerial images is a challenging task due to the variable sizes of the vehicles (small, medium, and large), high/low density of vehicles, and complex background in the camera's field of view. Therefore, it is important to design and develop robust vehicle detection algorithms suitable for aerial scenes.

The literature for vehicle detection in aerial images can be divided into descriptor-based and feature learning-based methods. The traditional feature descriptor-based approaches generally consist of three stages: vehicle localization, feature extraction, and classification. For vehicle localization, sliding window [1]–[5] is one of the most widely used methods. However, parameter selection for window size, stride length, etc., influences the detection performance. It also increases the processing time, which is not aligned with the demand for real-time detection. Therefore, many segmentation approaches, such as simple linear iterative clustering [6], edge-weighted centroidal Voronoi tessellations-based algorithm [7], have been proposed to alleviate some of these shortcomings. For feature extraction and classification, various feature descriptors, i.e., Haar-features, scale invariant feature transform, local binary pattern, histogram of oriented gradients (HOG), Gabor filters, etc., [8]–[12], have been used with support vector machines (SVM) for classification in the literature.

Liu and Mattyus [1] detected the vehicle locations by a sliding window mechanism using integral channel features (ICFs) and an AdaBoost classifier in a soft-cascade structure. The detected spatial regions are further classified into different orientations and vehicle type based on HOG features. Similar framework with a different set of descriptors to encode the local distributions of gradients, colors, and texture were proposed in [2]. Xu *et al.* [3] improved upon the original Viola–Jones object detection scheme for better vehicle detections in UAV images. Zhaou *et al.* [4] proposed to use a bag of words, local steering kernel descriptor, and orientation aware scanning mechanism to perform vehicle detection. More recently, Wu *et al.* [5] presented an aerial object detection framework, integrating diverse channel features extraction, feature learning, fast image pyramid matching, and boosting strategy.

Chen *et al.* [10] and [11] proposed the superpixel segmentation technique along with fast sparse representation to generate relevant vehicle patches. The HOG features for these patches are used in an SVM classifier for vehicle detection. Yu *et al.* [12] performed rotation-invariant object detection using superpixel segmentation and Hough forests.

The feature learning-based methods have utilized convolutional neural network (CNN) to learn features from an image for object detection. These methods can be categorized into two-stage and single-stage frameworks. Recent approaches [13]–[17] for aerial images have primarily used the two-stage architecture [fast/faster region-based CNN (R-CNN) [18]]-based frameworks to detect vehicles in aerial scenes. The faster R-CNN, consists of a region proposal network (RPN) and object detection network, leading to significant computational cost. Redmon *et al.* [19] proposed a unified one-stage model named YOLO to perform object detection and classification. Further, improvements were also proposed through recent detectors such as YOLOv2 [20], YOLOv3 [21], and RetinaNet [22]. However, these techniques are more suitable for images captured from canonical views and consist of a large number of parameters requiring high memory space. In addition, to effectively deal with the challenges of rotation variations and appearance ambiguity in geospatial scenes, various rotation-invariant

Manuscript received September 25, 2018; revised January 15, 2019 and May 10, 2019; accepted June 1, 2019. This work was supported by the Department of Science and Technology, Goverment of India, through the Science and Engineering Research Board under Project #SERB/F/9507/2017. (*Corresponding author: Santosh Kumar Vipparthi.*)

The authors are with the Vision Intelligence Lab, Department of Computer Science and Engineering, Malaviya National Institute of Technology, Jaipur 302017, India (e-mail: murarimandal.cv@gmail.com; manalshah3112@gmail.com; prashant23meena@gmail.com; sanhitadevi00@gmail.com; skvipparthi@mnit.ac.in).

Color versions of one or more of the figures in this letter are available online at http://ieeexplore.ieee.org.

Digital Object Identifier 10.1109/LGRS.2019.2923564

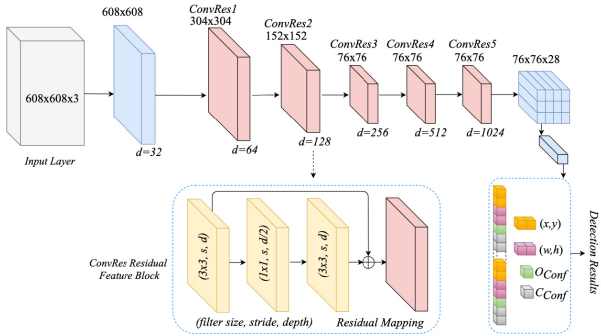

Fig. 1. Proposed AVDNet vehicle detection framework (for two vehicle classes). $x$, $y$, $w$, $h$: bounding box coordinates (center location, width, height), $O_{Conf}$: object confidence, $C_{Conf}$: class confidence.

detectors [8], [23]–[25] have been proposed in the literature. Diao et al. [26] proposed saliency-based object detection using deep belief networks. Nie et al. [15] used multitask models to combine semantic labeling and detection information for more accurate detection results.

In this letter, we address the shortcomings of the existing deep networks for object detection in aerial scenes and propose the one-stage vehicle detection network (AVDNet) object detector. The AVDNet preserves the small object features by introducing *ConvRes* blocks at multiple scales. To detect the densely populated objects, the AVDNet generates enlarged feature maps in the final layer of the network. The input layer is enlarged to maintain higher pixels-per-object values. Furthermore, we proposed a recurrent-feature aware visualization (RFAV) technique to visually analyze the AVDNet layers. The AVDNet also offers superior resource (computation and memory space) efficiency as compared to the state-of-the-art techniques.

## II. PROPOSED METHOD

We propose a novel AVDNet technique for aerial scenes. The detailed description of the proposed work, the motivation behind the methods and analysis of AVDNet is given in Sections II-A and II-B.

### A. AVDNet Object Detector

We designed a one-stage AVDNet object detector to simultaneously perform object localization and classification. The proposed detector uses the AVDNet convolutional network to learn the salient feature maps from the input image. The AVDNet generates a fixed-size tensor (76 × 76), which contains the bounding-box coordinates, object, and class confidence values of different anchor boxes for an input image. These features are then used to perform object localization and classification. The entire architecture of the proposed aerial object detector is shown in Fig. 1.

### B. AVDNet Convolutional Network

The aerial images usually consist of smaller and crowded objects, which makes it is very difficult to learn the individual object features. Typically, the initial CNN layers contain detailed information as compared to the more abstract features available in the deeper layers. Therefore, we propose to use residual feature blocks at multiple scales to preserve the low-level features present in the shallower layers while increasing the depth of the network. The proposed AVDNet consists of two convolutional (*conv*) layers and five residual feature blocks (*ConvRes*), as shown in Fig. 1. Each *ConvRes* block extracts the salient features by applying two 3 × 3 and one 1 × 1 *conv* operation. These *ConvRes* blocks enhance the capability of neurons to learn the minute details while maintaining the robustness of the features, as shown in Fig. 2. All the *conv* layers are followed by a batch normalization and leaky ReLu activation layer. The detailed explanation of the architecture is given as follows.

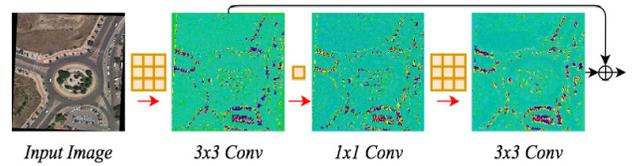

Fig. 2. RFAV visualization of the ConvRes3 block of AVDNet.

*1) Convolutional Layer:* Let $I_C(a, b)$ be an input image of size $M \times M$; $a \in [1, M], b \in [1, M]$ having $C$ channels and $f(\cdot)$ is the filter with a kernel size $h \times h$. The response of the convolutional layer (*conv*) is computed by the following equation:

$$F^d = \sum_{j=1}^{C} f^k(h) * I_j^n + b^k \bigg|_{k=1}^{d} \quad (1)$$

where $b^k$ is the bias, $n \in [1, M]$, and $d$ is the filter depth. In AVDNet, *conv* with stride 2 is used to downsample the feature maps, which allows inherent learning of the weights to represent the salient features from the previous layer. The outcome of *conv* with stride 2 is computed by the following equation:

$$F_{M/2}^d = \sum_{j=1}^{d} f^k(h) * I_j^{2n-1} + b^k \quad n \in [1, M/2]. \quad (2)$$

*2) ConvRes Blocks:* The response of a *ConvRes* block consisting of three *conv* layers is computed using the following equation:

$$F_{\text{ConvRes}}^d = F_l^d(a, b) + F_{l-2}^d(a, b) \quad (3)$$

where $l$ is the current *conv* feature layer. These *ConvRes* residual features are studied at three different scales in the AVDNet. The 1 × 1 *conv* response along with the leaky ReLu introduces an increased amount of nonlinearity in the feature response of the previous 3 × 3 *conv* layer. This enhances the ability of the network to study the local features for very small objects.

### C. Analysis of the AVDNet Detector

*1) Recurrent-Feature Aware Visualization:* In Fig. 2, the composite visual representation of the multiple feature maps generated at the end of a *conv* operation is shown. For $d$ feature maps in *conv* layer $l$, the RFAV representation is computed using the following equation:

$$\text{RFAV}_l(a, b) = \arg\max_z \left( H_l^{(a,b)}(z) \right); z \in [0, 255] \quad (4)$$

where $\arg\max(\cdot)$ collects the histogram bin index of the maximum value. The temporal histogram $H_l^{(a,b)}(\cdot)$ at pixel

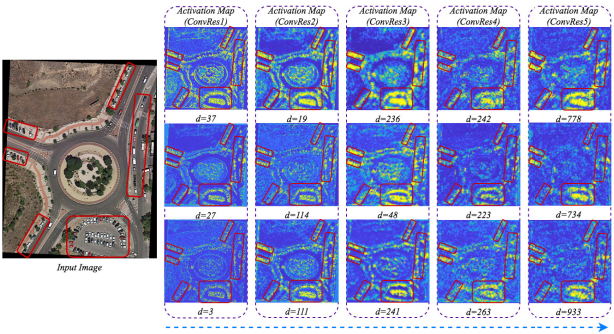

Fig. 3. Sample activation responses after each ConvRes block of AVDNet. The red boxes highlight the activations in different regions for the presence of vehicles in the input image, $d$ = depth of the activation map.

location $(a, b)$ is calculated using the following equation:

$$H_l^{(a,b)}(z) = \sum_{k=1}^{d} \delta\big(F_l^k(a, b) - z\big); z \in [0, 255]. \quad (5)$$

*2) Feature Degradation Problem:* Usually, the initial layers have detailed information as compared to the features at the deeper layers. These detailed features are very useful in the detection of small and dense objects. In order to preserve the small-sized object features, we designed residual feature blocks at multiple scales in the AVDNet. These residual connections enable the network to activate the regions with the presence of vehicles and neglect the rest, as shown in Fig. 3.

*3) Effect of Final Feature Map Resolution:* The lower pixels-per-object values of the smaller objects cause the features to vanish in the deeper networks. This is in contrast to the features of bigger objects with higher pixel-per-object values, which are clearly detected in the deeper CNN layers. For example, if the resolution of an image is 1024 × 1024, and the size of the vehicle is ∼40 × 60 in the image, then, in the YOLOv2 architecture, the image is first resized into 416 × 416 or 608 × 608 input size and further downsampled five times after several convolution and max-pooling layers. This will result in negligible feature representation (around 1 or 2 tensors) for the dense small-sized vehicles in the output feature maps, which is insufficient to accurately detect those objects. Therefore, the proposed AVDNet addresses this issue by maintaining a high-dimensional tensor shape in the final layer.

*4) Higher Pixels-Per-Object Values:* We have made another observation that the input layer size influences the network's capability to learn the features for the small-size objects. This point was also reiterated by Lin *et al.* [22] in RetinaNet where they used 600-pixel and 800-pixel image scale as input to the network to improve the detection performance. In the AVDNet, we have used an enlarged input layer size of 608 × 608. However, the total number of parameters and inference model size of the AVDNet is approximately 79% and 63% lower as compared to YOLO and RetinaNet, respectively. The proposed AVDNet is more accurately able to perform vehicle detection with lower space and computational complexities.

## III. EXPERIMENTAL RESULTS AND DISCUSSION

### A. Data Sets

*1) VEDAI:* VEDAI [27] data set contains aerial images captured from various scenarios for vehicle detection. In our experiments, we have trained our proposed AVDNet for 11 vehicle categories. The details of all the data sets (number of images, objects per class, etc.) are given in Table I.

TABLE I
SUMMARIZATION OF THE EVALUATED DATA SETS

| Dataset | #Images | #Objects | #Object per class |
|---------|---------|----------|-------------------|
| VEDAI | 1248 | 3773 | car: 1393, truck: 307, pickup: 955, tct: 190, cc: 397, bt: 171, mc: 4, bus: 3, van: 101, other: 204, large: 48 |
| DLR-3K | 262 | 8401 | car: 8210, hv: 191 |
| DOTA | 1558 | 55235 | car: 24516, hv: 11307, pln: 4733, bt: 14679 |
| ABD | 79 | 1396 | car: 1353, hv: 11, bt: 32 |
| Complete | 3099 | 68579 | car: 36510, hv: 12406, pln: 4781, bt: 14882 |

*tct: tractor, cc: camping car, mc: motorcycle, hv: heavy vehicle, pln: plane, bt:boat

*2) DLR-3K:* DLR-3K [1] is mainly comprised of scenes from urban and residential areas. For our experiments, we have divided each image (total of 20 images) into 16 parts to generate 320 images. We have manually annotated all the images in DLR-3K and generated 8401 horizontally aligned bounding boxes for all the objects. Finally, we selected 262 images with a resolution of 1404 × 936 for training and evaluation.

*3) DOTA:* DOTA [28] introduced a large-scale data set consisting of 2806 aerial images. In our experiments, we have represented these objects through four categories: car, heavy vehicle, plane, and boat. Moreover, we manually annotated all the images in DOTA and generated 55 235 horizontally aligned bounding boxes as ground truth.

*4) Airborne Data Set (ABD) Data Set:* We collected 79 new aerial images from online sources and generated a new data set named ABD by annotating 1396 objects for our experiments. The objects were annotated with four different classes: car, heavy vehicle, plane, and boat.

*5) Complete Data Set:* For more comprehensive performance analysis of the proposed and existing object detectors in aerial scenes, we generated a large data set by combining VEDAI, DLR-3K, DOTA, and ABD data sets. The complete data set is categorized into four classes similar to the DOTA and ABD data set. The summary description of all the data sets is given in Table I.

### B. Experimental Settings

*1) Implementation Details:* The entire method is implemented in darknet. The detection results of the AVDNet depend on various parameters, such as intersection over union (IoU) thresholds and number of anchor boxes. The threshold is the minimum object confidence score for which the network will detect an object. The object and class confidence values are computed, as given in [20]. We have generated four anchors with respect to each of the four training data sets.

*2) Training Configuration:* Training is done over a Titan Xp GPU system with stochastic gradient descent optimizer and minibatch size = 4. The weight decay and momentum parameters are set to 0.0005 and 0.9, respectively. The training loss is calculated by taking the sum of square error from the final layer of the network, as given in [20]. We train our model with input layer size of 608 × 608, which is determined through a set of experiments on parameter sensitivity analysis for computational performance and model accuracy. Similarly,

TABLE II
COMPARATIVE DETECTION PERFORMANCE OF THE AVDNET AND EXISTING STATE-OF-THE-ART TECHNIQUES

| Method/mAP (%) | VEDAI | DLR-3K | DOTA | Complete |
|---|---|---|---|---|
| Coupled R-CNN | 12.04 | 11.74 | 25.60 | 19.66 |
| YOLOv2_416x416 | 9.08 | 9.61 | 33.36 | 28.86 |
| YOLOv2_608x608 | 25.12 | 26.81 | 47.45 | 48.04 |
| Faster R-CNN | 34.82 | 20.04 | 42.29 | 38.02 |
| YOLOv3_416x416 | 32.07 | 52.11 | 74.46 | 70.35 |
| YOLOv3_608x608 | 38.98 | 54.49 | 76.60 | 75.21 |
| RetinaNet | 43.47 | 54.77 | 73.77 | 71.28 |
| **AVDNet** | **51.95** | **56.24** | **79.65** | **80.02** |

the initial learning is set to 0.001 through experimental analysis with different learning rates.

*3) Model Training:* We divide VEDAI, DOTA, Complete data set into train and test set with a ratio of ∼[90:10]. The DLR-3K data set is divided with a ratio of ∼[80:20]. The AVDNet is trained over each data set without using any pretrained weights. The AVDNet detector is trained for ∼30k iterations over VEDAI, DOTA, Complete data set, and ∼15k iterations over DLR-3K. The initial learning rate is further reduced by a factor of 10 at 20k iterations for VEDAI, DOTA, and Complete data set, and at 10k iterations for DLR-3K. The RetinaNet detector was trained for each data set using the ResNet-50 model pretrained over the ImageNet data set. Similarly, pretrained ResNet-101 was used while training Faster R-CNN over the aerial data sets. The AVDNet generates four bounding boxes corresponding to each grid cell as a center and selects the bounding box with the highest IoU with respect to the given threshold.

### C. Results and Analysis

*1) Quantitative Results:* The performance measures of the proposed AVDNet and other state-of-the-art approaches for vehicle detection in VEDAI, DLR-3K, DOTA, and Complete data set is given in Table II. We compare different methods in terms of mAP, which corresponds to the average of the maximum precisions at different recall values. To ensure fair comparison, all the methods were evaluated over the same set of unseen test data.

The proposed AVDNet outperforms the existing state-of-the-art techniques in all four aerial data sets. More specifically, it achieves 8.48%, 1.47%, 5.88%, and 8.74% higher mAP in comparison to RetinaNet over VEDAI, DLR, DOTA, and Complete data set respectively. Similarly, it outperforms YOLOv3 by 12.97%, 1.75%, 3.05%, and 4.81% over the four respective data sets, respectively. The AVDNet also significantly exceeds the performance of Coupled R-CNN [16], which is designed for the task of aerial object detection.

We present the precision-recall graphs at different IoU thresholds for AVDNet, YOLOv2, and YOLOv3 (Figs. 4 and 5). It is evident from Figs. 4 and 5 and Table II, that the AVDNet is much more robust for vehicle detection in aerial images in comparison to the current popular CNN-based techniques. We have calculated the results at two different input layer size 416 and 608, to evaluate the performance of YOLOv2. The performance improvement of YOLOv2_608x608 over YOLOv2_416x416 clearly proves our analysis about the effects of pixel-per-object values over the performance, as discussed earlier in Section III. The proposed AVDNet outperforms both YOLOv2_416x416 and

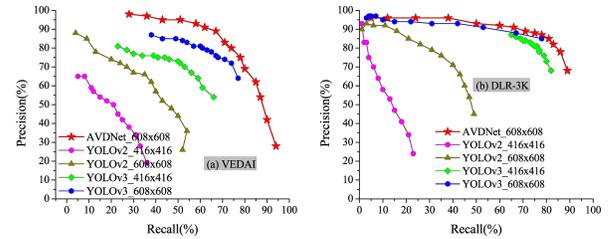

Fig. 4. Precision-recall graph of the proposed and existing state-of-the-art object detectors over (a) VEDAI and (b) DLR-3K.

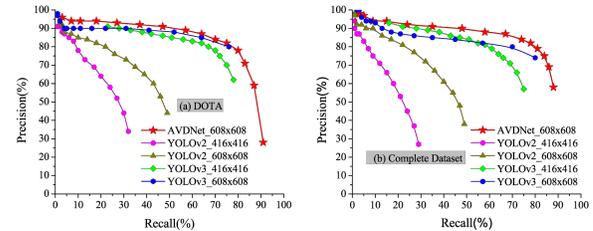

Fig. 5. Precision-recall graph of the proposed and existing state-of-the-art object detectors over (a) DOTA and (b) Complete data set.

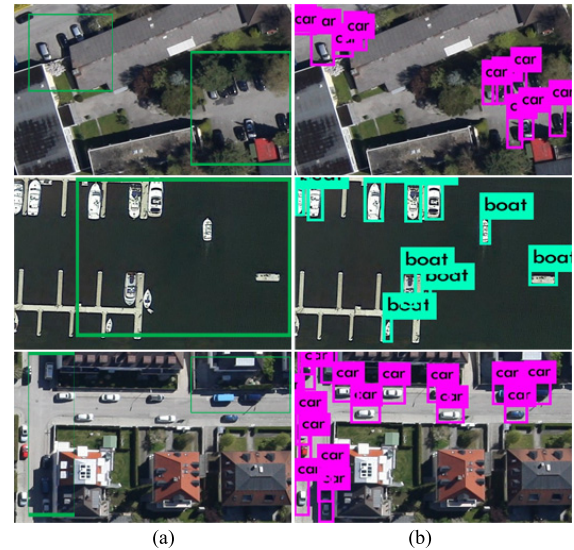

Fig. 6. Qualitative performance of AVDNet under various challenging scenarios. First row: occlusion by overhead building/trees. Second row: vehicles of varying sizes and orientations. Third row: vehicles covered with shadows. (a) Input aerial image. (b) Vehicles detected by AVDNet.

YOLOv2_608x608 in terms of mAP. As stated earlier, the enlarged dimensionality of the final tensor layer used in the proposed AVDNet leads to better performance as compared to YOLOv2 and YOLOv3, and the same is shown in Figs. 4 and 5.

*2) Qualitative Results:* We show the qualitative results of our approach to different challenging scenarios in Fig. 6. The detection responses from the original images are cropped out for appropriate visual representation. The AVDNet is able to detect vehicles, which are partially occluded by overhead building or trees, as shown in the first row in Fig. 6. Similarly, the varying shapes and orientation of vehicles (boats) are also robustly detected by the proposed method, and the same is shown in the second row in Fig. 6. Furthermore, vehicles covered with shadows incident from buildings/trees/other objects are also detected accurately by the proposed AVDNet.

TABLE III
COMPUTATIONAL AND SPACE COMPLEXITY OF THE PROPOSED METHOD AND EXISTING STATE-OF-THE-ART TECHNIQUES

| Method | No. of params. (in millions) | Model size |
| --- | --- | --- |
| YOLOv2 | 67 | 255 MB |
| YOLOv3 | 61 | 235 MB |
| Faster R-CNN | 59 | 253 MB |
| RetinaNet | 36 | 146 MB |
| **AVDNet** | **13** | **53 MB** |

We can also see that, in case of a high degree of occlusion, the vehicles are undetected as shown in the fourth row as a failure case. However, overall, these qualitative results demonstrate the effectiveness of our approach in different challenging scenarios in aerial object detection.

*3) Complexity Analysis:* The computation and space complexity of the proposed method and existing state-of-the-art techniques is given in Table III. We can see that the proposed method uses approximately 1/5, 2/9, 5/14 times smaller number of parameters as compared to YOLO (v2 and v3), Faster R-CNN, and RetinaNet, respectively. Similarly, the proposed AVDNet model utilizes approximately 2/9, 2/9, 5/14 times less memory space as compared to YOLO (v2 and v3), Faster R-CNN, and RetinaNet, respectively. Thus, the proposed method offers superior resource efficiency (computation and memory space) as compared to the state-of-the-art techniques.

## IV. CONCLUSION

In this letter, we identified the shortcomings of the existing one-stage object detectors for aerial scenes. To address these issues, we proposed a new object detector AVDNet by introducing *ConvRes* blocks at multiple scales to preserve salient features of small-sized objects. We maintained higher pixels-per-object values and generated enlarged feature maps for accurate feature representation in the output layer. Moreover, we proposed to analyze the network behavior by introducing RFAV technique. Furthermore, we generated a new data set ABD by collecting 79 new aerial images (annotated 1396 objects) from open sources. We demonstrated the efficacy of the AVDNet by conducting experiments on three challenging data sets VEDAI, DLR-3K, and DOTA. We also developed a large-scale aerial image data set by combining all three data sets along with the ABD data set. From the experimental results, it is clear that the AVDNet outperforms the existing state-of-the-art approaches in terms of mAP, computational (no. of parameters), and space complexity. The detection performances over the complete data set also provide baseline results for future vehicle detection techniques designed for aerial scenes.